# Improve Academic Query Resolution through BERT-based Question Extraction from Images

Nidhi Kamal
*data science lab*
*embibe*
Bengaluru, India
nidhi.k@embibe.com

Saurabh Yadav
*data science lab*
*embibe*
Bengaluru, India
saurabh.yadav@embibe.com

Jorawar Singh
*data science lab*
*embibe*
Bengaluru, India
jorawar.singh@embibe.com

Aditi Avasthi
*data science lab*
*embibe*
Bengaluru, Country
aditi@embibe.com

*Abstract*—Providing fast and accurate resolution to the student's query is an essential solution provided by Edtech organizations. This is generally provided with a chat-bot like interface to enable students to ask their doubts easily. One preferred format for student queries is images, as it allows students to capture and post questions without typing complex equations and information. However, this format also presents difficulties, as images may contain multiple questions or textual noise that lowers the accuracy of existing single-query answering solutions. In this paper, we propose a method for extracting questions from text/images using a BERT-based deep learning model and compare it to the other rule-based and layout-based methods. Our method aims to improve the accuracy and efficiency of student query resolution in Edtech organizations.

## I. Introduction

Providing fast and accurate resolution to student queries is a critical goal of the online education system. These queries are often submitted through a chatbot-like interface, which allows students to easily ask their doubts. However, for the system to correctly answer the query, it must be clear and contain a single question. In practice, academic queries may contain complex equations, tables, images, or other relevant information, which can be difficult for students to communicate through text alone. One solution to this problem is for students to upload an image of their query, which eliminates the need for them to type out complex equations and information. However, this approach presents its own set of challenges.

Images may contain multiple questions or extra textual noise, which can lower the accuracy of existing single query answering solutions and decrease user satisfaction. This can be attributed to the extra effort required from students to crop clear and concise images and the lack of understanding of the intricacies and sensitivity of question answering services to noise or the student's intention to ask multiple questions. Traditional approaches to addressing this problem include providing guidelines to students on how to crop their images to include only a single question or using rule-based text splitting techniques to identify individual questions within an image. However, these methods have proven to be ineffective, with around 30% of images submitted still containing textual noise or multiple questions. Additionally, rule-based approaches introduce the problem of identifying the correct splitting criteria and often fail to generalize to new cases.

Deep learning-based models, such as BERT and LayoutLMv3 have shown to be highly effective in capturing contextual information. BERT, a transformer-based model, is pre-trained on a large corpus of text and fine-tuned for specific tasks such as NER. It has been shown to be highly effective in capturing contextual information and achieving state-of-the-art performance on various NER benchmarks. Similarly, LayoutLM is a transformer-based model that is pre-trained on a large corpus of images and fine-tuned for specific tasks such as information extraction from images. In this paper, we propose to use these models for extracting questions from text and images based queries. We compare these techniques based on accuracy and effort required in adoption of these techniques. Finally, the paper presents BERT-based question extraction from both images and complex text queries as the best-performing deep learning approach for improving student query resolution.

## II. Related Work

Parsing questions from the student query can sometimes be challenging. If the query is not parsed correctly, the student query may remain unanswered. The traditional approach of question parsing is to use deterministic parsers. Typically, these parsers are learned on manually designed syntactic patterns. Slav et. al. proposed an approach where they have proposed the POS tag based dependency parsing [1]. However, the rule based approaches are domain specific and tend to fail for complex question parsing where the context of the question also needs to be parsed.

There has been pre-training and fine-tuning of the BERT based models for multiple natural language tasks. Taher et. al. [5] presents an approach of fine-tuning the BERT model for Persian language's entity detection task. There are works where authors [6] have pre-trained the BERT specifically for the NER task by using Wikipedia to create a huge NER corpus. The author claims to outperform BERT on NER across 9 domains. However, the available work in BERT-NER closely match our problem statement. We needed to identify and extract the only entity (user question) from the text, and we were unable to find prior work available for question parsing using language models.

LayoutLM achieves state-of-the-art results where the task is both text and image centric. One such scenario is to extract the questions from images. There has been work on information extraction from documents using architecture based on LayoutLM [2]. Prior to LayoutLMv3, there were works where authors proposed architectures to facilitate cross modal learning for example DocFormer [3] and SelfDoc [4]. Research work has shown that layout dependent extraction tasks from documents are best addressed by multimodal architectures. But the only problem is to collect the labeled dataset. For the multimodal transformers, the bounding box coordinates are required along with the BIO tags for the input sequence. It is a very difficult and time-consuming process to augment the data for fine-tuning multimodal transformers.

## III. THE PROPOSED APPROACH

In our proposed method for extracting multiple questions from an image input, we first use an Optical Character Recognition (OCR) service to extract the raw text from the image. Then, we use a token classification task, specifically Named Entity Recognition (NER), to identify the question spans within the raw text. The text is tokenized using a tokenizer corresponding to the BERT or LayoutLM model, and each word token is labeled with one of the following BIO tags:

- B-Question denotes the beginning token of a question
- I-Question denotes the intermediate token of a question
- O denotes other token

After applying tokenizer, if a word is broken into subtokens, we adjust the BIO tags by replicating the parent token's tag for the subtokens. However, if the parent token's tag is B-Question, we consider its subtoken's tag as I-Question. An example of the word-level tokens and their corresponding BIO tags is:

Answer the following. What is force?
O           O   O       B   I

We are fine-tuning the pretrained BERT on our dataset for the multiple question detection. For the evaluation of LayoutLM model on multiple question extraction, we follow a similar token classification approach, where both raw text and images are provided as inputs. To evaluate the rule based approach, regex based rule system adopted from our existing system. This tries to remove text noise by identifying question starts and question end patterns.

We fine-tuned the model for 20 epochs with SGD optimizer, CrossEntropy loss function and 2e-5 learning rate.

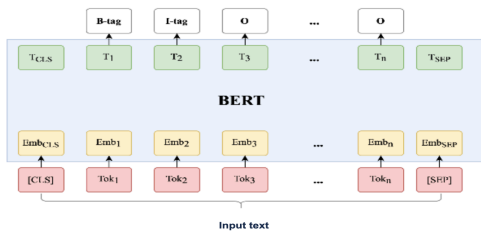

Fig. 1.  Architecture of the multiple question identification model..

BERT based model was supplied with extra augmented data. Data was augmented by randomly appending the noise text with no questions before and after the question entities and addition of random question entities from other data points. These augmentations increased the sample size to ~22k. This type of augmentation was not possible for LayoutLM as it required the natural bending of images and changes in bounding boxes for image domain input.

## IV. EXPERIMENTS AND RESULTS

### A. Dataset

In our study, we have a dataset of approximately 300 tagged images. These images were used for training and evaluating our proposed method for extracting multiple questions from an image input. We separated 53 images from the dataset for use in the validation set, which were used to evaluate the performance of all three approaches.

For BERT training, we were able to augment and increase the number of data points by using training data point along with some additional text that contained either a single question or no question at all. Through this augmentation process, we were able to generate approximately 22,000 data points for training. For LayoutLM this was not done as it also required image domain manipulations.

### B. Evaluation Method

Model accuracy performance was evaluated on standard precision and recall metrics on the validation dataset. The metrics are calculated as:

- Precision: Ratio of the number of correctly identified questions to the total number of predicted questions by the model
- Recall: Ratio of the number of correctly identified questions to the actual number of questions

Compute performance of models was evaluated as average time taken per query (excluding OCR time) in milliseconds on a 6 core Intel Xeon processor. Model size is measured in terms of number of parameters.

### C. Results and Comparison of Pretrained Models

*a)* Table I compares the accuracy of BERT and LayoutLM approaches to rule-based approach with precision and recall metrics. Inference time performance is compared in Table 2 in terms of parameter count (model size representing approximate memory requirements) and time taken per query. BERT based model outperformed the other approaches in terms of accuracy while being significantly smaller and faster than the layoutLM model.

TABLE I.     MODEL ACCURACY COMPARISON

| S.No. | Model | Precision | Recall |
|---|---|---|---|
| 1 | Rule Based Approach | 53 | 30 |
| 2 | LayoutLMv3 | 90 | 61 |
| 3 | BERT | 96 | 83 |

a.     Benchmarks of different approaches

TABLE II.     PERFORMANCE RESULTS

| S.No. | Model Architecture | CPU Time (6 cores) | Parameter Count |
|---|---|---|---|
| 1 | Rule Based Approach | 2.5 ms | NA |
| 2 | LayoutLMv3 | 526 ms | 133 M |
| 3 | BERT | 205 ms | 107 M |

b.     Performance

*D. Qualitative Analysis and Discussion*

a) BERT based model successfully extracts questions from raw text without image input while being significantly smaller and faster than layoutLM model. Table III showcases the sample of input images and extracted questions. It is also more suitable for adoption in the question answer pipeline at scale. Some advantages include easy augmentation of training data, less compute requirements and latency in response. LayoutLM is also more restrictive in license for commercial adoption, which forces users to adopt older versions.

TABLE III. EXAMPLES

| Column 1 |
|---|
| 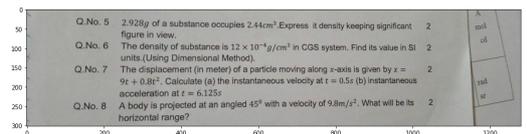 |
| **1st question** |
| No. $5 \quad 2.928 \mathrm{~g}$ of a substance occupies $2.44 \mathrm{~cm}^{3}$.Express it density keeping significant 2 figure in view. |
| **2nd question** |
| The density of substance is $12 \times 10^{-4} \mathrm{~g} / \mathrm{cm}^{3}$ in CGS system. Find its value in SI 2 units.(Using Dimensional Method). |
| **3rd question** |
| The displacement (in meter) of a particle moving along $x$-axis is given by $x=2$ $9 t+0.8 t^{2}$. Calculate (a) the instantaneous velocity at $t=0.5 \mathrm{~s}$ (b) instantaneous acceleration at $t=6.125 \mathrm{~s}$ |
| **4th question** |
| A body is projected at an angled $45^{\circ}$ with a velocity of $9.8 \mathrm{~m} / \mathrm{s}^{2}$. What will be its 2 horizontal range? |
| 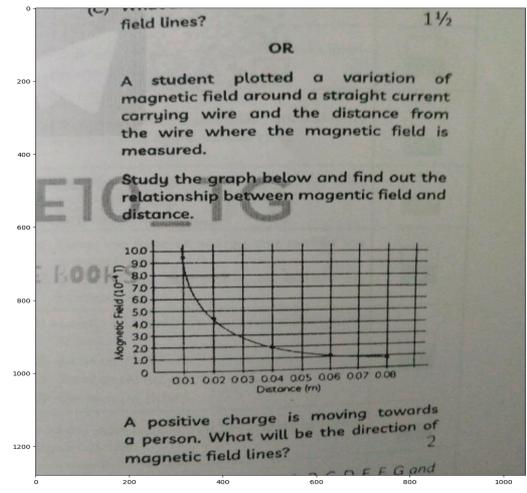 |
| **Question 1** |
| A student plotted a variation of magnetic field around a straight current carrying wire and the distance from the wire where the magnetic field is measured. Study the graph below and find out the relationship between magentic field and distance. |
| **Question 2** |
| A positive charge is moving towards a person. What will be the direction of magnetic field lines? |
| 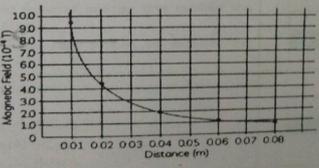 |
| **1st question** |
| 1There are 2 pillars which are cylindrical in shape. Also 2 domes at the corners which are hemispherical. 7 smaller domes at the centre. Flag hoisting ceremony on Independence Day takes place near these domes. 2(i) How much cloth material will be required to cover 2 big domes each of radius $2.5$ metres? (Take $\pi=22 / 7$ ) 3(a) $75 \mathrm{~m}^{2}$ 4(b) $78.57 \mathrm{~m}^{2}$ 5(c) $87.47 \mathrm{~m}^{2}$ 6(d) $25.8 \mathrm{~m}^{2}$ |
| **2nd question** |
| 1(ii) Write the formula to find the volume of a cylindrical pillar. 2(a) $\pi r^{2} \mathrm{~h}$ 3(b) $\pi r l$ 4(c) $\pi r(l+r)$ 5(d) $2 \pi r$ |
| **3rd question** |
| (iii) Find the lateral surface area of two pillars if height of the pillar is $7 \mathrm{~m}$ and radius of the base is $1$ |
| **4th question** |
| 1(v) What is the ratio of sum of volumes of two hemispheres of radius $1 \mathrm{~cm}$ each to the volume of a sphere of radius $2 \mathrm{~cm}$ ? 2(a) $1: 1$ 3(b) $1: 8$ 4(c) $8: 1$ 5(d) $1: 16$ |

c. Column 1 represents the input image, and corresponding questions detected by our model.

## V. CONCLUSION

This paper presents a deep learning based approach to extract the questions from the text or image query from a limited dataset. The method uses a pre trained BERT model for token classification, with the goal of identifying the question spans within the raw text extracted from the image. We compared this method with rule based and LayoutLM based methodologies and found that BERT model outperformed the other methods in terms of accuracy and efficiency. The method is easier to finetune, simple to use, and supports good augmentation to expand training data. Our proposed approach provides significant improvement from traditional approaches with precision of 96% and recall 83%.

In future the model can easily be extended to other languages such as Indic languages. This approach has major dependency on OCR performance on image query. OCR free transformer models can also be evaluated for improving this task. LayoutLM based models can also be further evaluated with more sophisticated augmentation strategies and larger datasets to improve question extractions from more complex layouts.

## REFERENCES

[1]. Slav Petrov, Pi-Chuan Chang, Michael Ringgaard, and Hiyan Alshawi, "Uptraining for accurate deterministic question parsing," In Proceedings of the 2010 Conference on Empirical Methods in Natural Language Processing, pages 705–713, Cambridge, MA. Association for Computational Linguistics, 2010.

[2]. Hong, T., Kim, D., Ji, M., Hwang, W., Nam, D., & Park, S., "BROS: a pre-trained language model focusing on text and layout for better key

information extraction from documents," AAAI Conference on Artificial Intelligence, 2021.

[3]. Appalaraju, S., Jasani, B., Kota, B.U., Xie, Y., & Manmatha, R., "DocFormer: end-to-end transformer for document understanding," IEEE/CVF International Conference on Computer Vision (ICCV), 973-983, 2021.

[4]. Li, P., Gu, J., Kuen, J., Morariu, V.I., Zhao, H., Jain, R., Manjunatha, V., & Liu, H., "SelfDoc: self-supervised document representation learning," IEEE/CVF Conference on Computer Vision and Pattern Recognition (CVPR), 5648-5656, 2021.

[5]. Taher, E., Hoseini, S.A., & Shamsfard, M., "Beheshti-NER: Persian named entity recognition using BERT," ArXiv, abs/2003.08875, 2020.

[6]. Liu, Z., Jiang, F., Hu, Y., Shi, C., & Fung, P., "NER-BERT: a pre-trained model for low-resource entity tagging," ArXiv, abs/2112.00405, 2021.

[7]. Cheung, Z., Phan, K.L., Mahidadia, A., Hoffmann, A., "Feature extraction for learning to classify questions," In: Webb, G.I., Yu, X. (eds) AI 2004: Advances in Artificial Intelligence. AI 2004. Lecture Notes in Computer Science(), vol 3339. Springer, Berlin, Heidelberg, 2004.

[8]. Mikhail Arkhipov, Maria Trofimova, Yuri Kuratov, and Alexey Sorokin, "Tuning multilingual transformers for language-specific named entity recognition," In Proceedings of the 7th Workshop on Balto-Slavic Natural Language Processing, pages 89–93, Florence, Italy. Association for Computational Linguistics, 2019.

[9]. Yang X, Zhang H, He X, Bian J, Wu Y, "Extracting family history of patients from clinical narratives. Exploring an end-to-end solution with deep learning models," JMIR Med Inform 2020;8(12):e22982, 2019.

[10]. Huang, Y., Lv, T., Cui, L., Lu, Y., & Wei, F., "LayoutLMv3: pre-training for document AI with unified text and image masking," Proceedings of the 30th ACM International Conference on Multimedia, 2022.

[11]. Devlin, J., Chang, M., Lee, K., & Toutanova, K., "BERT: pre-training of deep bidirectional transformers for language understanding," ArXiv, abs/1810.04805, 2019.